\documentclass[final,5p,times,twocolumn]{elsarticle}

\usepackage{amssymb}
\usepackage{subfigure}
\usepackage{multicol}
\usepackage{multirow}

\usepackage{amsmath}
\usepackage{caption}

\usepackage[table]{xcolor}
\definecolor{mygray}{gray}{.9}

\usepackage[pagebackref=true,breaklinks=true,colorlinks,citecolor=citecolor,bookmarks=false]{hyperref}

\newlength\savewidth

\usepackage{booktabs}

\makeatletter
\newcommand{\thickhline}{%
	\noalign {\ifnum 0=`}\fi \hrule height 1pt
	\futurelet \reserved@a \@xhline
}
\makeatother

\usepackage{newfloat}
\usepackage{listings}
\journal{Neurocomputing}

\begin{document}

\begin{frontmatter}



\title{TIVE: A Toolbox for Identifying Video Instance Segmentation Errors}


%\author[1]{Wenhe Jia}
%\ead{jiawh@bupt.edu.cn}

%\author[1]{Lu Yang}
%\ead{soeaver@bupt.edu.cn}

%\author[1]{Zilong Jia}
%\ead{jzl@bupt.edu.cn}

%\author[1]{Wenyi Zhao}
%\ead{2013211876@bupt.edu.cn}

%\author[1]{Yilin Zhou}
%\ead{ylzhou@bupt.edu.cn}

%\author[1]{Qing Song\corref{mycorrespondingauthor}}
%\ead{priv@bupt.edu.cn}

%\affiliation[1]{
%	organization={Beijing University of Posts and Telecommunications, Artificial Intelligence Academy},%Department and Organization
%	addressline={10th xitucheng road, Haidian District}, 
%	city={Beijing},
%	postcode={100086}, 
%	state={Beijing},
%	country={China}
%}

\author{Wenhe Jia}
\ead{jiawh@bupt.edu.cn}

\author{Lu Yang}
\ead{soeaver@bupt.edu.cn}

\author{Zilong Jia}
\ead{jzl@bupt.edu.cn}

\author{Wenyi Zhao}
\ead{2013211876@bupt.edu.cn}

\author{Yilin Zhou}
\ead{ylzhou@bupt.edu.cn}

\author{Qing Song\corref{mycorrespondingauthor}}
\cortext[mycorrespondingauthor]{Corresponding author}
\ead{priv@bupt.edu.cn}

\affiliation{
	organization={Beijing University of Posts and Telecommunications, Artificial Intelligence Academy},%Department and Organization
	addressline={10th xitucheng road, Haidian District}, 
	city={Beijing},
	postcode={100086}, 
	state={Beijing},
	country={China}
}

%%%%%%%%%%% Abstract %%%%%%%%%%%
\begin{abstract}	
Since first proposed, Video Instance Segmentation(VIS) task has attracted vast researchers' focus on architecture modeling to boost performance. Though great advances achieved in online and offline paradigms, there are still insufficient means to identify model errors and distinguish discrepancies between methods, as well approaches that correctly reflect models' performance in recognizing object instances of various temporal lengths remain barely available. More importantly, as the fundamental model abilities demanded by the task, spatial segmentation and temporal association are still understudied in both evaluation and interaction mechanisms. 

In this paper, we introduce TIVE, a \textbf{T}oolbox for \textbf{I}dentifying \textbf{V}ideo instance segmentation \textbf{E}rrors. By directly operating output prediction files, TIVE defines isolated error types and weights each type's damage to mAP, for the purpose of distinguishing model characters. By decomposing localization quality in spatial-temporal dimensions, model's potential drawbacks on spatial segmentation and temporal association can be revealed. TIVE can also report mAP over instance temporal length for real applications. We conduct extensive experiments by the toolbox to further illustrate how spatial segmentation and temporal association affect each other. We expect the analysis of TIVE can give the researchers more insights, guiding the community to promote more meaningful explorations for video instance segmentation. The proposed toolbox is available at \url{https://github.com/wenhe-jia/TIVE}.
\end{abstract}


%%%%%%%%%%% Highlights %%%%%%%%%%%
%\item TIVE bin false positives and false negatives produced by video instance segmentor into 7 types, including general object recognition errors and two spatial-temporal localization errors with specific focus on spatial segmentation and temporal association quality of predicted mask sequence.

%\item By directly operating output prediction files, TIVE can isolate error predictions and weight each type’s damage to mAP, in purpose of distinguishing model characters. To evaluate models' practicality in real scenario, TIVE can report mAP of different video instance tenporal ranges.

%\item We select several state-of-the-art VIS models and investigate the relation of spatial and temporal localization abilities with TIVE, which are ideally to benefit from each other. Through extensive experiments, we found that most of current methods cannot promote spatial segmentation and temporal association together, they only focus on at most one of two aspects, this phenomenon need further study by the community.


\end{frontmatter}


%%%%%%%%%%% sec 1 %%%%%%%%%%%
\section{Intruduction}
\label{Intruduction}
As the indispensable technique in numerous real applications, \emph{e.g.}, video surveillance and editing, autonomous driving, Video Instance Segmentation (VIS)\cite{Youtubevis,Mots,Ovis,Bdd100k,Surveyforvideosegmentation} is emerging among various vision tasks\cite{Airnet,Cpc,Vcpsurvey,Faceattack,Lbsi,Longtailsurvey} in recent years. Compared to image-level instance segmentation, video instance segmentors are required to assign unique identities to video instances. Demand for both spatial segmentation and temporal association leaves VIS at the intersection of mask-level object recognition and sequence modeling, making it one of the most fundamental role among video object recognition tasks(\emph{e.g.}, Multi-Object Tracking (MOT)\cite{Motchallenge}, Video Object Segmentation (VOS)\cite{Davis,Youtubevos} and Video Instance Parsing (VIP)\cite{Vip}). However, discussion about error sources, model abilities, and interacting mechanism on the above-mentioned aspects is unavailable, as well proper evaluating scheme for recognizing video instances with different attributes.

% figure 1 ++++++++++
\begin{figure}[!t]
	\centering
	\includegraphics[width=1.0\linewidth]{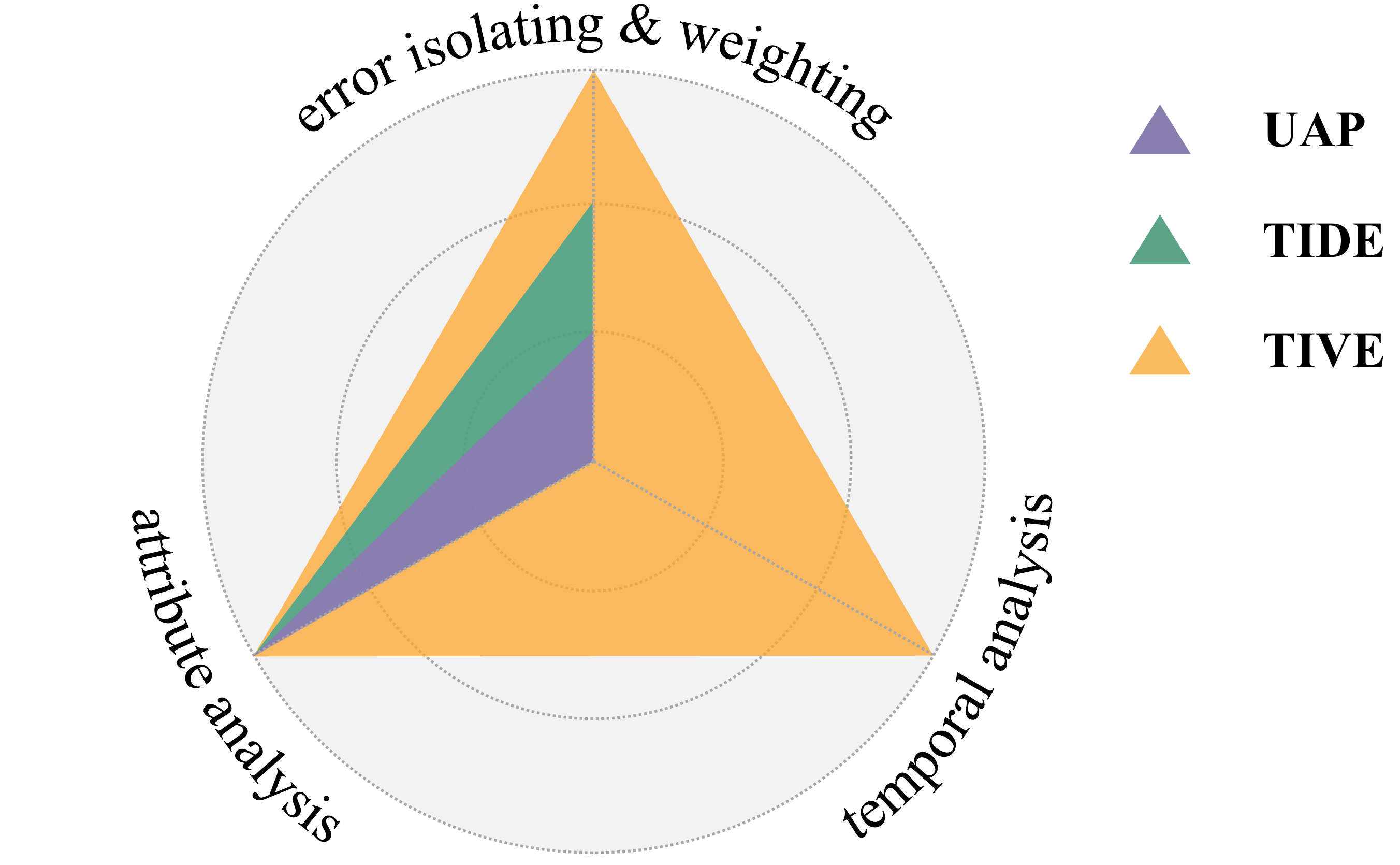}
	\captionsetup{font={normalsize}, justification=raggedright}
	\caption{\textbf{Comparison between TIVE and other error analyzing toolboxes.} UAP\cite{Uap} weight error contribution in a progressively fixing mechanism, while TIDE\cite{Tide} can give objective and isolated error analysis. But both of them cannot distinguish spatial and temporal misslocalized false positives.}
	\label{fig:ToolboxComparison}
	\vspace{-4.5mm}
\end{figure}
%+++++++++++++++++++

Though continuous works are proposed by leaps and bounds, the following problems still puzzle the community: firstly, we don't know how false positive and negative predictions relate to the overall metric. When optimizing mAP alone, we may inevitably neglect the relative importance of different types of errors that vary among applications, leaving discrepancies between algorithms unclear. For example, temporal association is crucial to recognize objects that disappear or are occluded temporarily in video surveillance, and accurate spatial segmentation is required by autonomous driving systems to precisely formulate obstacle avoidance operations; secondly, no appropriate scheme to evaluate model performance over instance temporal length. For attribute analysis, performances in different temporal lengths are notably important, but the official evaluation scheme is not based on instance temporal length, but on the length of videos; lastly, the relation between spatial segmentation and temporal association is not investigated. As the most required abilities, spatial segmentation and temporal association are expected to promote mutually, thus can leave a positive impact on another when promoting one. Unfortunately, there is few work to discuss the relation between them.

There are some existing error analyzing toolboxes that may partially solve the above-mentioned problems. Several image-level error analyzing toolboxes\cite{Diagnosing,Uap,Tide} try to diagnose errors and observe how much they contribute to the performance decline, but they fail to distinguish errors distributed in spatial and temporal dimensions. Some video-level tools\cite{Tracklinic,Diagnosingvideoralationerrors, Diagnosingactiondetectionerrors} focus on diagnosing video action and relation errors, not video objects. They pay attention to the subjective impacts brought by annotators and task-specific object attributes(\emph{e.g.}, context size for action, relation distribution, \emph{etc.}), which are not applicable for VIS, not to mention demonstrating the relation of spatial segmentation and temporal association.

Thus we introduce TIVE, a novel \textbf{T}oolbox for \textbf{I}dentify various \textbf{V}ideo instance segmentation \textbf{E}rrors. Decomposing general localization error in spatial and temporal dimensions, TIVE clearly subdivides 7 isolated error types, as well as explores model performance on different instance temporal lengths. By weighting the error contributions to mAP damage by individually fixing oracles, we can understand how these error sources relate to the overall metric, which is crucial for algorithm development and model selection in deployment. The variation of spatial segmentation and temporal association error weights can laterally reflect the model ability change. Evaluating performance over instance temporal length can help the community to evaluate models for real scenarios. Figure.\ref{fig:ToolboxComparison} shows the comparison between TIVE and other error analyzing toolboxes.

Providing comprehensive analysis of several typical algorithms, clear discrepancies between methods are revealed by error weights, we find that short video instances that live less than 16 frames are harder to recognize for all methods. Only one of the investigated algorithms can enable spatial segmentation and temporal association to benefit from each other, while others generally meet at most one aspect, this phenomenon may demand further exploration by the community. Due to the modulated functional design, we can easily extend TIVE to other video object recognition tasks, \emph{e.g.}, MOT\cite{Motchallenge,Visdronemot21}, VOS\cite{Davis,Youtubevos,Zeroshotvos,Lavrsscl,Rgar,Hop,Wsre,Iaa,Siampolar,Ldr,Seemore,Mcgfae,Tocmsa} and VIP\cite{Vip} task, whose metric calculation have strong similarity with video instance segmentation, and the principle of TIVE is also referential to identify errors in Video Semantic Segmentation(VSS)\cite{Cityscapes,Vspw} and Video Panoptic Segmentation(VPS)\cite{Vps} task.

%%%%%%%%%%% sec 2 %%%%%%%%%%
\section{Related Work}
%%%%% sec 2.1 %%%%%
\subsection{Video instance segmentation}
As an extension of image-level instance segmentation task\cite{Maskrcnn,Parsingrcnn,Pdrnet,Cpssgnet,Mvsn,Sunnet,Rprcnn,Tprparsing,Solov2}, current video instance segmentation methods can be roughly divided into online and offline paradigms, which derive from MOT, VOS and newly raised vision transformer techniques.

\vspace{1mm}
\textbf{Online methods} select one frame as reference and one or several other frames as query, where ground  truth labels and masks of query frames are considered as learning targets\cite{Youtubevis,Sipmask,Stmask,Sgnet,Visolo}. At inference stage, they first perform frame-level instance segmentation by object detector\cite{Fasterrcnn,Fpn,Fcos,Cpmrcnn,Heirrcnn} or instance segmentor\cite{Maskrcnn,Solov2,Yolact}, then conduct temporal association with tracking modules\cite{Centertrack,Fairmot,Trackformer}, which is usually conducted under manual-designed rules and representation comparison. Except for the pioneer Mask Track R-CNN\cite{Youtubevis}, later works tend to leverage more frame-level predictions to refine results of each query frame\cite{Visolo}, \emph{e.g.}, classification scores and predicted masks, which provide rich temporal references.

\vspace{1mm}
\textbf{Offline methods} take several randomly sampled frames from a video clip as input both in training and inference progress and directly predict mask sequences, labels and masks from all sampled frames are supervision signals. Maskprop\cite{Maskprop} and Proposereduce\cite{Proposereduce} combine mask propagation technique from VOS tasks with frame-level instance segmentation models to segment video instances in spatial-temporal dimensions. Specifically, they use Mask R-CNN\cite{Maskrcnn} to get frame-level instance categories and masks, then propagate them to the entire video clip. Compared to the propagation-based methods that have a complicated processing pipeline to generate sequence results for multiple video instances, the transformer-based methods dominate the state-of-the-art performance\cite{Vistr,Ifc,Seqformer,Mask2former,Tevit} recently. Thanks to the strong ability to capture global context, this type of models directly learn to segment mask sequences during training and produce sequence-level predictions in only one-time inference.

%%%%% sec 2.2 %%%%%
\subsection{Error analyzing tools}
Although previous literature provides qualitative proofs to demonstrate their model superiorities over others, but limited visual comparisons are incomplete and nonobjective. There exists toolboxes identifying relative vision recognition errors in frame and video level may provide useful guidance.

\vspace{1mm}
\textbf{Image-level toolboxes.} UAP\cite{Uap} tried to explain the effects of object detection errors based on cocoapi, subjective error types and fixing oracles are defined to explore the metric upper bounds. But with progressive weighting scheme, it fails to isolate contributions of errors.  TIDE\cite{Tide} is the most recent image-level object recognition error analyzing toolbox, which clearly defines isolated errors and weighting contribution of each by individually fixing oracles, providing meaningful observations and suggestions to mainstream methods and algorithm design.

% figure 2 ++++++++++
\begin{figure*}[!t]%
	\centering
	\subfigure[Ground truth.]{
		\label{fig:Groundtruth}
		\begin{minipage}[t]{1.0\linewidth}
			\centering
			\includegraphics[width=1.0\linewidth]{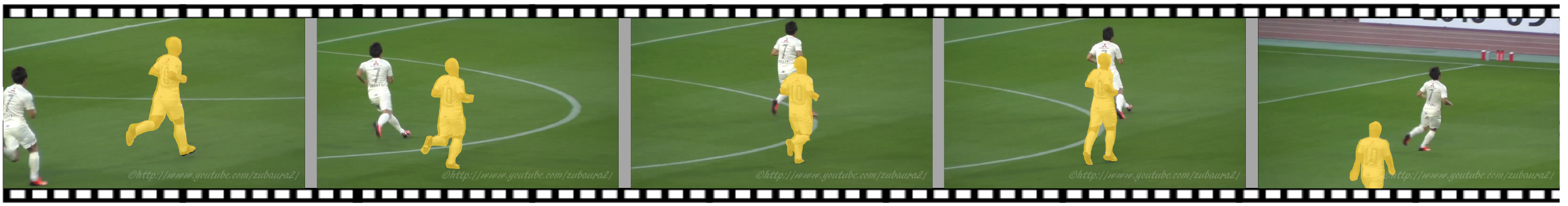}
		\end{minipage}
	}
	\subfigure[Spatial segmentation error.]{
		\label{fig:SpatError}
		\begin{minipage}[t]{1.0\linewidth}
			\centering
			\includegraphics[width=1.0\linewidth]{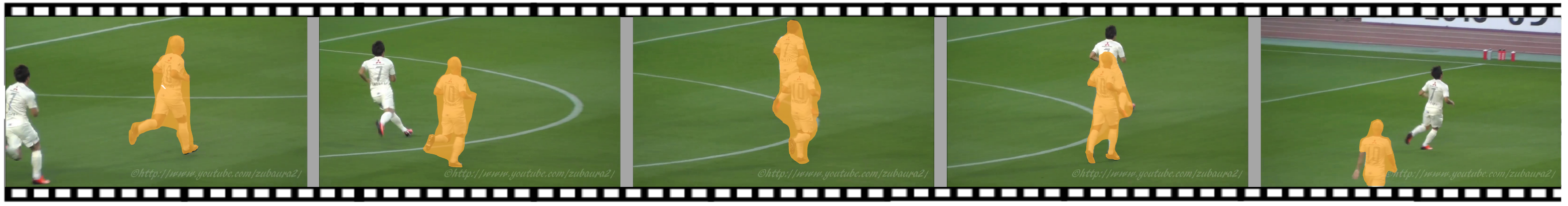}
		\end{minipage}
	}
	\subfigure[Temporal association error.]{
	\label{fig:TempError}
	\begin{minipage}[t]{1.0\linewidth}
		\centering
		\includegraphics[width=1.0\linewidth]{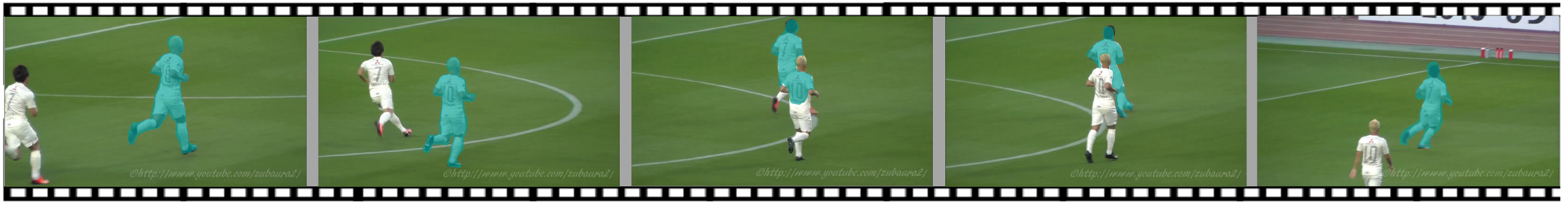}
	\end{minipage}
	}
	\captionsetup{font={normalsize}, justification=raggedright}
	\caption{\textbf{Complicated cases of localization error for video instance segmentation.} Under mask sequence $IoU$ threshold of 0.5, localization error of the same video instance can appear in different circumstances. (a) Ground truth. (b) and (c) are spatial miss-segmented prediction and temporal miss-associated prediction with mask sequence $IoU$ = 0.48 respectively.
	}
	\label{fig:LocalizationError}
\end{figure*}%
%+++++++++++++++++++

\vspace{1mm}
\textbf{Video-level toolboxes.} Few related works search for identifying video recognition errors, they more focus on 1) exploring challenging factors for object tracking based on self-established dataset, whose instances distribute in no more than one challenge factor\cite{Tracklinic}, thus keeping models from handling complicated data distribution; 2) diagnosing detection errors for human actions and video relations, rather than focusing on video objects. Chen \emph{et.al.} \cite{Diagnosingvideoralationerrors} analyzed the subjective factors of annotations and gave some conclusions about effects of them, while Alwassel \emph{et.al.} \cite{Diagnosingactiondetectionerrors} studied the sources of video relation detection errors and explore the metric sensitivities to each.

\vspace{1mm}
Although the community has been improving the model performance by the only judgment of mAP, toolbox to explore model discrepancies remains barely available, so we introduce TIVE to fill up this vacancy.

%%%%%%%%%%% sec 3 %%%%%%%%%%%
\section{TIVE}
In this section, we first discuss the limitations of the official evaluation scheme(\S\ref{sec:RevisitMetric}), then demonstrate our toolbox design with objective errors(\S\ref{sec:ErrorDesign}), at last we take further analysis on attribute of instance temporal length(\S\ref{sec:RangeAnalysis}).

%%%%% sec 3.1 %%%%%
\subsection{Revisiting evaluation metric}
\label{sec:RevisitMetric}
\vspace{1mm}
\textbf{Calculating mAP.} Since newly raised, current VIS algorithms just evaluate their models on the most popular benchmark, YouTube-VIS, which adopts a variation of cocoapi as its evaluation implementation\footnote{https://github.com/youtubevos/cocoapi}. Given a video clip with $T$ frames , mask sequence intersection-over-union(IoU) is calculated between a ground truth mask sequence $\mathbf{m}_{p...q}^i$ and a predicted mask sequence $\mathbf{\tilde{m}}_{\tilde{p}...\tilde{q}}^i$ by Eq.\ref{eq:SeqIoU}:

\begin{equation}
	\centering
	\label{eq:SeqIoU}
	\textit{IoU} = \frac{\Sigma_{t=1}^T|\mathbf{m}_t\cap\mathbf{\tilde{m}}_t|}{\Sigma_{t=1}^T{|\mathbf{m}_t\cup\mathbf{\tilde{m}}_t|}}
\end{equation}
where $\mathbf{m}_{t}$ and $\mathbf{\tilde{m}}_{t}$ denote annotated and predicted binary segmentation mask of each frame, with $p\in[1,T]$, $\tilde{p}\in[1,T]$ and $q\in[p,T]$, $\tilde{q}\in[\tilde{p},T]$ denote their starting and ending time respectively. For instances appear in the middle of a video clip, $p$, $\tilde{p}$ are extended to 1, $q$ and $\tilde{q}$ are entented to T by padding empty masks during evaluation.

\vspace{1mm}
After the matching manipulation, standard average precision($AP$) and recall($AR$) are calculated by multiple IoU thresholds same as COCO evaluation. The localization quality of a prediction is considered as the aggregation of spatial segmentation and temporal association quality, therefore making the evaluation progress concise and elegant. 

% figure 3 ++++++++++
\begin{figure*}[!t]
	\centering
	\includegraphics[width=1.0\linewidth]{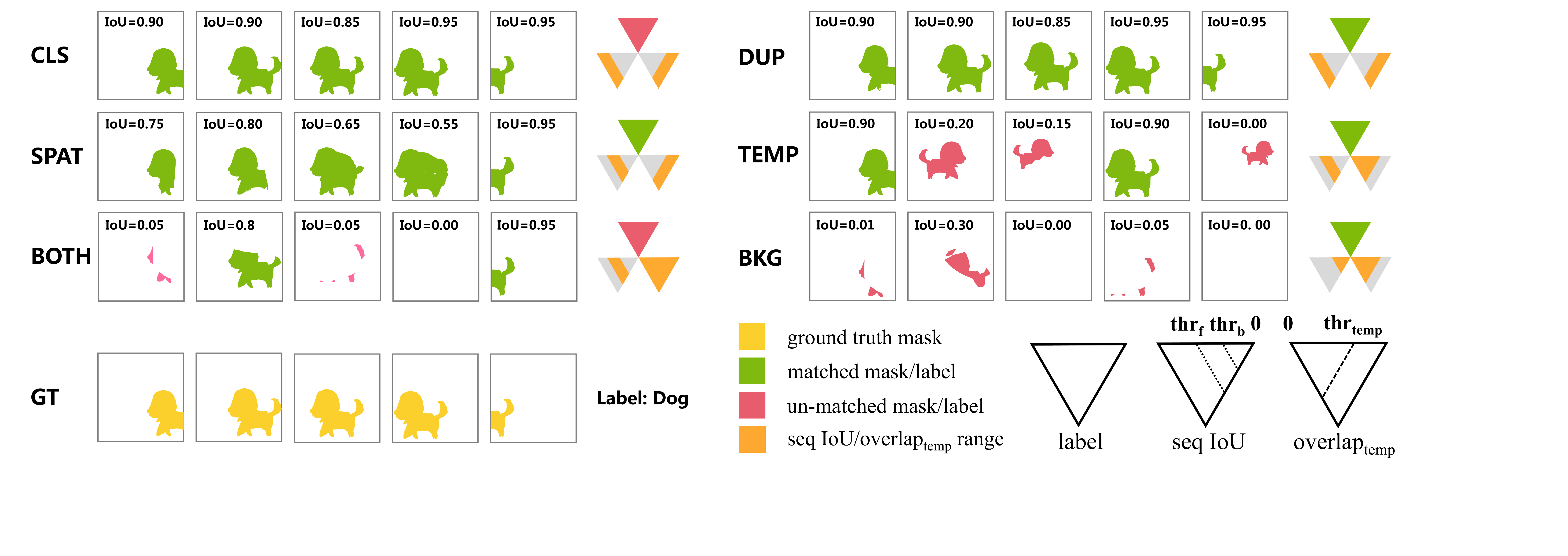}
	\captionsetup{font={normalsize}, justification=raggedright}
	\caption{\textbf{Error type definitions.} We define 7 types of errors for video instance segmentation, predicted category, mask sequence IoU, and temporal overlap of each error type are shown in the \textit{right rows} of images, the missed ground truh error is not shown.}
	\label{fig:ErrorType}
\end{figure*}
%+++++++++++++++++++

\vspace{1mm}
\textbf{Limitation of evaluation toolkit.} Following the official implementation, we may get puzzled in identifying different false positives that fail to accurately localize video instances. Look at Figure.\ref{fig:Groundtruth}, considering the football player in the central area of the first frame, the miss-localized prediction with the right category could be in various cases. Figure.\ref{fig:SpatError} suffers from sub-optimal spatial segmentation while Figure.\ref{fig:TempError} contains an identity switch between two human instances. Improvements in model ability from either aspect may fix these errors to true positives, but without a large amount of visualization, researchers can not get a picture of their models' potential shortness to localize instances in spatial-temporal dimensions.

\vspace{1mm}
The official implemented mask sequence IoU confuses the model's localization performance of two dimensions together, resulting in no clear illustration of algorithm characters. With the additional temporal dimension, recognizing instance sequences in videos becomes more complicated and challenging. So an error analyzing toolbox is eagerly demanded, for the purpose of distinguishing model discrepancy, as well as giving suggestions on model design.

%%%%% 3.2 %%%%%
\subsection{Error definition}
\label{sec:ErrorDesign}
Errors produced by models can be roughly divided into false positives and false negatives. Still, these two error types mix up the factors referenced during evaluation(\emph{e.g.}, mask category and location). We bin all error predictions into 7 types as Figure.\ref{fig:ErrorType}(\emph{w/o.} missed ground truth error, which usually represents false negative predictions).

\vspace{1mm}
\textbf{General recognition errors.} We first use $\textit{IoU}_{max}$ to denote a false positive’s mask sequence IoU with its best-matched ground truth of the given category, the foreground IoU threshold $\textit{thr}_{f}$ and the background threshold $\textit{thr}_{b}$ are set to 0.5 and 0.1 unless otherwise noted. By two variables of category and $\textit{IoU}_{max}$, 6 error types come along as described in \cite{Tide}: classification error(\textbf{Cls}), duplication error(\textbf{Dup}), localization error(\textbf{Loc}), both classification and localization error(\textbf{Both}), background error(\textbf{Bkg}) and missed ground truth error(\textbf{Miss}). These general errors that derive from the $AP$ calculation are applicable for both image-level and video-level object recognition tasks.

\vspace{1mm}
\textbf{Spatio-temporal localization errors.} After the general error types, we delve into possibilities to distinguish general localization errors into more specific sub-categories. TIVE performs a reverse analysis relative to the IoU formulation: decomposes the localization quality into spatial and temporal sub-assessments, representing spatial segmentation and temporal association qualities, respectively.

Here we introduce temporal overlap $\textit{overlap}_{temp}$ to represent the tracking quality of predictions. Suppose a predicted mask sequence and a ground truth mask sequence as described in \S\ref{sec:RevisitMetric}, we first compute image-level mask IoUs $\{\textit{IoU}_{t}\}^T_{t=1}$ of all frames by Eq.\ref{eq:FrameIoUs}:

\begin{equation}
	\centering
	\label{eq:FrameIoUs}
	\{\textit{IoU}_{t}\}^T_{t=1} = \{\frac{|\mathbf{m}_t\cap\mathbf{\tilde{m}}_t|}{|\mathbf{m}_t\cup\mathbf{\tilde{m}}_t|}\}^T_{t=1}
\end{equation}
where $\textit{IoU}_{t}$ represents mask IoU at each frame; then we count the matched frame number $\textit{N}_{match}$ whose $\textit{IoU}_{t}$ are higher than frame mask IoU threshold $\textit{thr}_{spat}$(0.1 as default); finally the $\textit{overlap}_{temp}$ is the ratio of $\textit{N}_{match}$ to their temporal union as Eq.\ref{eq:TemporalOverlap}:

\begin{equation}
	\centering
	\label{eq:TemporalOverlap}
	\textit{overlap}_{temp} =  \frac{\mathbf{N}_{match}}{\{p...q\}\cup\{\tilde{p}...\tilde{q}\}}
\end{equation}
where $\{p...q\}$ and $\{\tilde{p}...\tilde{q}\} \in [1,T]$ represent frame index sets for non-empty masks.

With $\textit{IoU}_{max}$ and $\textit{thr}_{temp}$(0.7 as default), localization error with mask sequence IoU between ($\textit{thr}_{b}$, $\textit{thr}_{f}$) can be distinguished into novel spatial segmentation error and temporal association error are as bellow:

\begin{itemize}
	\item \textbf{Spatial Segmentation Error(Spat)}: $\textit{thr}_{b} \leq \textit{IoU}_{max} \leq \textit{thr}_{f}$ for GT of the {\it correct}
	class (i.e., classified correctly but localized incorrectly), meanwhile with $\textit{overlap}_{temp} \geq \textit{thr}_{temp}$.
	
	\item \textbf{Temporal Association Error(Temp)}: $\textit{thr}_{b} \leq \textit{IoU}_{max} \leq \textit{thr}_{f}$ for GT of the {\it correct}
	class (i.e., classified correctly but localized incorrectly), meanwhile with $\textit{overlap}_{temp} \leq \textit{thr}_{temp}$.
\end{itemize}

% figure 4 ++++++++++
\begin{figure*}[!t]
	\label{fig:MethodErrorDistribution}
	\centering
	\subfigure[Error distribution cross models.]{
		\label{fig:ErrorDist}
		\begin{minipage}[t]{1.0\linewidth}
			\centering
			\includegraphics[width=1.0\linewidth]{figs/ModelSummaryOnYTVISMiniVal.pdf}
		\end{minipage}
	}
	\subfigure[Online models.]{
		\label{fig:OnlineModels}
		\begin{minipage}[t]{0.48\linewidth}
			\centering
			\includegraphics[width=1.0\linewidth]{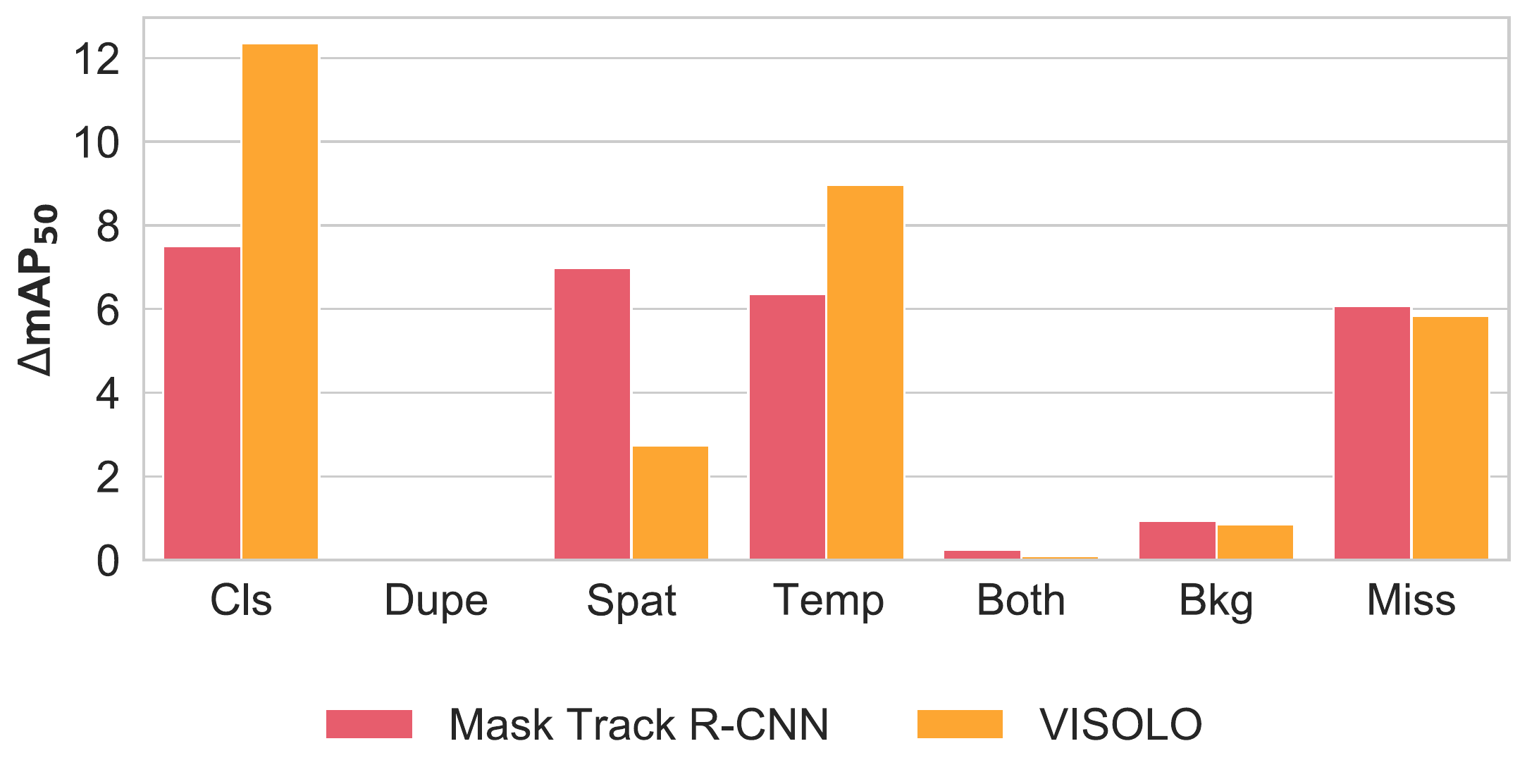}
		\end{minipage}
	}
	\subfigure[Offline models.]{
		\label{fig:OfflineModels}
		\begin{minipage}[t]{0.48\linewidth}
			\centering
			\includegraphics[width=1.0\linewidth]{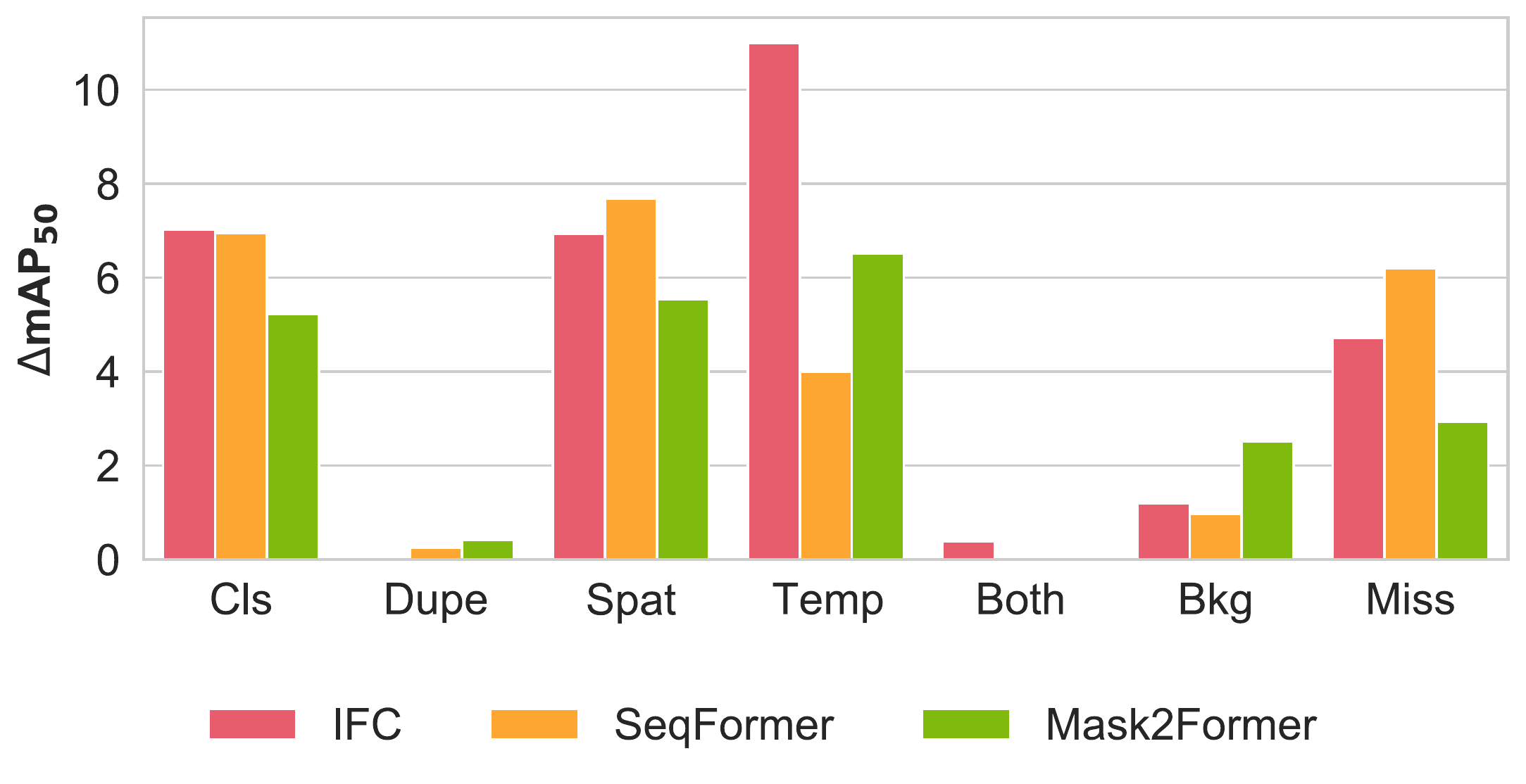}
		\end{minipage}
	}
	\captionsetup{font={normalsize}, justification=raggedright}
	\caption{\textbf{Comparison cross video instance segmentation methods.} Models are all trained on YouTube-VIS-2021 mini-train subset, and error distributions are reported on mini-val subset. We report mean metrics and error weights of 5 runs for VISOLO, IFC, and Mask2Former. The following tables and charts are the same.}
\end{figure*}
%+++++++++++++++++++

When calculating metrics, there are no intermediate states to evaluate the effects of fixing errors progressively, as the fixing oracle of a false positive or false negative prediction will change the matching distribution. So progressive fixing scheme cannot isolate error types, which may mismeasure their contributions to performance decline. Motivated to eliminate confusion, the effects of each error type are measured by observing variation $\Delta AP@50$ after individually fixing type errors as described in \cite{Tide}, note that the sum of AP and all error weights is not equal to 100 AP, but we can get 100 AP after fixing all errors.

%%%%% 3.3 %%%%%
\subsection{Analysis over temporal ranges}

\label{sec:RangeAnalysis}
As the mostly used benchmark, most original videos of YouTube-VIS 2019 and 2021 have a length of around 3 to 6 seconds(15 to 30 frames), which is relatively not long enough to judge models' ability for real application. In the YouTube-VIS 2022 challenge, the sponsor released an additional set of long videos, aiming at observing the performance of models over long temporal range. For evaluation, the metric for the original YouTube-VIS 2021 validation and test subset are named mAP$_S$, and mAP$_L$ for additional long videos are evaluated with the official toolkit separately, formulation of the final mAP is averaged over metrics calculated on short and long videos.

% table 1 ++++++++++
\begin{table}[!t]
	\centering
	\captionsetup{font={normalsize}, justification=raggedright}
	\caption{Data distribution of YouTube-VIS-2021 train subset and our self-split mini train and validation subset.}
	%%\vspace{1mm}
	\setlength\tabcolsep{2.6mm}
	\renewcommand\arraystretch{1.4}
	\scalebox{1.0}{
		\begin{tabular}{l|c|cccc}
			\hline\thickhline
			\rowcolor{mygray}
			& & \multicolumn{4}{c}{Instances} \\
			\cline{3-6}
			\rowcolor{mygray}
			\multirow{-2}{*}{Subset} & \multirow{-2}{*}{Videos} & all & short & medium & long \\
			\hline
			\hline
			train & 2985 & 6283 & 843 & 3113 & 2327 \\
			mini-train & 2771 & 5804 & 782 & 2876 & 2146 \\
			mini-val & 214 & 479 & 61 & 237 & 181 \\
			\hline
	\end{tabular}}
	\label{table:YTVISDataDistrubution}
\end{table}
%+++++++++++++++++++

\vspace{1mm}
There exists two irrationalities: 1.) Firstly, rather than the length of original videos, the evaluation of model performance over long temporal range should be based on the length of instance sequences. because not all instances in the newly added long videos retain long lifetimes, and vice versa for instances in original videos; 2.) Secondly, averaging of mAP$_S$ and mAP$_L$ may leave an underlying vulnerability: participants can achieve better mAP by targeted optimizing mAP$_S$, because YouTube-VIS 2021 validation and test subset account for a larger proportion of all the videos and instances of YouTube-VIS 2022.

Length of video instances in YouTube-VIS 2021 training subset distribute centrally on 18$\sim$20, 28$\sim$31, and 34$\sim$36 frames, while others have much fewer samples as tail length. We define instances of temporal length between 0 and 16 as short, 16 to 32 as medium, 32 to 72 as long, which metrics named mAP$_s$, mAP$_m$, and mAP$_l$ respectively. 

% table 2 ++++++++++
\begin{table*}[!t]
	\centering
	\captionsetup{font={normalsize}, justification=raggedright}
	\caption{\textbf{Metrics and error weights cross temporal ranges for selected video instance segmentors on YouTube-VIS-2021 mini-val.} All models are trained on YouTube-VIS-2021 mini-train subset with ResNet-50 as backbone, the default training frame number is listed in the second column..}
	%%\vspace{3mm}
	\setlength\tabcolsep{1.6mm}
	\renewcommand\arraystretch{1.4}
	\scalebox{1.0}{
		\begin{tabular}{l|c|c|cccc|ccccccc}
			\hline\thickhline
			\rowcolor{mygray}
			& & & \multicolumn{4}{c|}{Metrics} & \multicolumn{7}{c}{Error Weights($\Delta$AP@50)} \\
			\cline{4-7}\cline{8-14}
			\rowcolor{mygray}
			\multirow{-2}{*}{Method} & \multirow{-2}{*}{Pretrain} & \multirow{-2}{*}{\shortstack{Training\\Frames}} & mAP & mAP$_s$ & mAP$_m$ & mAP$_l$ & Cls & Dup & Spat & Temp & Both & Bkg & Miss \\
			\hline
			\hline
			Mask Track R-CNN & MSCOCO & T=2 & 36.30 & 10.44 & 35.87 & 43.43 & 7.50 & 0.00 & 6.98 & 6.36 & 0.24 &  0.94 & 6.08 \\ 
			VISOLO  & MSCOCO & T=3 & 38.49 & 8.58& 39.31 & 40.94 & 12.35 & 0.00 & 2.74 & 8.97 & 0.09 & 0.86 & 5.84 \\
			\hline
			IFC & MSCOCO & T=5 & 40.27 & 20.44 & 38.98 & 40.98 & 7.03 & 0.02 & 6.93 & 10.99 & 0.38 & 1.20 & 4.72 \\
			SeqFormer & MSCOCO & T=5 & 43.37 & 21.69 & 40.43 & 39.20 & 6.95 & 0.25 & 7.69 & 4.00 & 0.00 & 0.97 & 6.20 \\
			Mask2Former & MSCOCO & T=2 & 48.07 & 25.05 & 43.85 & 44.37 & 5.23 & 0.42 & 5.54 & 6.52 & 0.06 & 2.51 & 2.94 \\
			\hline
		\end{tabular}
	}
	\label{table:MetricsOverTemporal}
\end{table*}
\vspace{5mm}
%+++++++++++++++++++

% figure 5 ++++++++++
\begin{figure*}[!t]
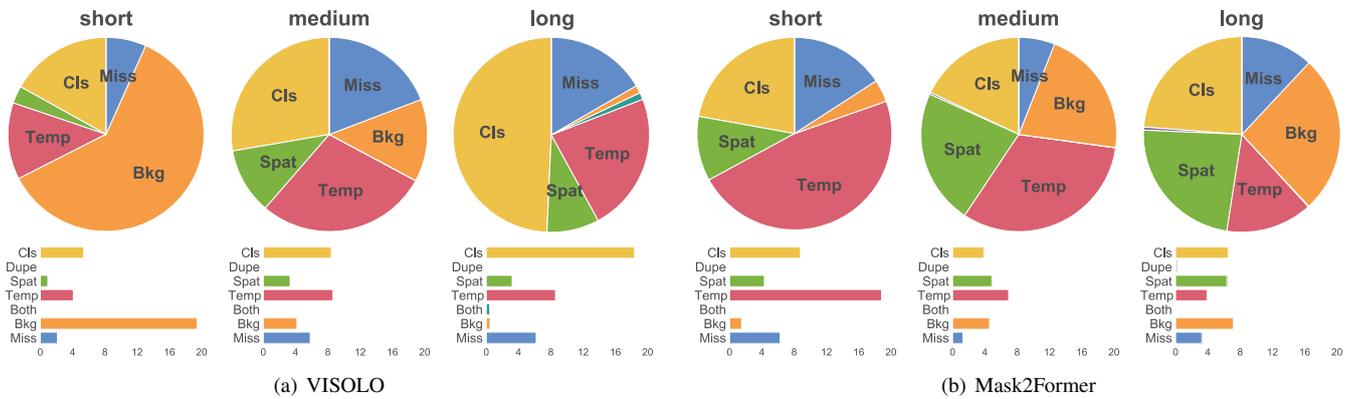

	\label{fig:ErrorDistOverTemporal}
	\centering
	\subfigure[VISOLO]{
		\label{fig:RangeErrorVISOLO}
		\begin{minipage}[t]{0.48\linewidth}
			\centering
			\includegraphics[width=1.0\linewidth]{figs/RangeErrorVISOLO.pdf}
		\end{minipage}
	}
	\subfigure[Mask2Former]{
		\label{fig:RangeErrorM2F}
		\begin{minipage}[t]{0.48\linewidth}
			\centering
			\includegraphics[width=1.0\linewidth]{figs/RangeErrorM2F.pdf}
		\end{minipage}
	}
	\captionsetup{font={normalsize}, justification=raggedright}
	\caption{\textbf{Error distribution over different video instance temporal ranges} for VILOLO and Mask2Former. Ground truth and predicted video instances beyond the specific temporal range are not considered for error weight calculation.}
	\vspace{-1mm}
\end{figure*}
%+++++++++++++++++++

\vspace{-4mm}
As the ground truth of YouTube-VIS validation and test subsets are not released, so we turn to the training subset for our study. We split the YouTube-VIS training subset into mini-train and mini-val subsets, both of them have a relatively balance category distribution \emph{w.r.t} the whole training subset, detailed quantity distribution of videos and instances is shown in Table.\ref{table:YTVISDataDistrubution}.

%%%%%%%%%%%  sec 4 %%%%%%%%%%%
\section{Experimental analysis}
In this section, we first use error distribution comparison cross models to verify our toolbox design(\S\ref{sec:VerifyDesign}); then, we will investigate models' performance over different temporal ranges, find out what errors fail the segmentors to accurately recognize video instances with specific temporal length(\S\ref{sec:AttributeAnalysis}); last, we conduct further studies on current methods to explore the relation of spatial segmentation and temporal association ability(\S\ref{sec:TemporalContext}) to see whether they can be promoted manually.

\vspace{1mm}
\textbf{Models.} We choose a series of open-source algirithms from both online and offline paradigms:
\vspace{1mm}
\begin{itemize}
	\item \textbf{Online methods.} We choose Mask Track R-CNN\cite{Youtubevis} and VISOLO\cite{Visolo} as the target methods. Mask Track R-CNN is the first algorithm developed for video instance segmentation, while VISOLO performs tracking based on grid similarity, achieving the best performance on YouTube-VIS-2021 among all the online methods.

	\item \textbf{Offline methods.} We choose IFC\cite{Ifc}, Seqformer\cite{Seqformer}, and Mask2Former\cite{Mask2former} due to their state-of-the-art performance. IFC and SeqFormer separately perform feature-level and query-level temporal aggregation, while Mask2Former is implemented with no interaction between temporal context during model forward.
\end{itemize}

\vspace{-2mm}
For fair comparison, all models are trained on YouTube-VIS-2021 mini-val subset with ResNet-50\cite{Resnet} as backbone. We report mean of 5 runs as the results for VISOLO, IFC, and Mask2Former due to their data sampling randomness.

%%%%% sec 4.1 %%%%%
\subsection{Verify toolbox design}
\label{sec:VerifyDesign}
In this subsection, we focus on analyzing error distributions to see whether TIVE's errors correctly reflect methods' characters to general intuition, summarizes as Figure.\ref{fig:ErrorDist}.

% table 3 ++++++++++
\begin{table*}[!t]
	\centering
	\captionsetup{font={normalsize}, justification=raggedright}
	\caption{\textbf{Performance cross models YouTube-VIS-2021 mini-val under different pretrain mode.} All models are trained by official release settings except for pretrain data and task. When pretrained with MSCOCO, stronger ability reflects in \textbf{SPAT} error weight decline.} 
	%%\vspace{3mm}
	\setlength\tabcolsep{2.8mm}
	\renewcommand\arraystretch{1.4}
	\scalebox{1.0}{
		\begin{tabular}{l|c|c|cccc|cccc}
			\hline\thickhline
			\rowcolor{mygray}
			& & & \multicolumn{4}{c|}{Metrics} & \multicolumn{4}{c}{Error Weights($\Delta$AP@50)} \\
			\cline{4-7}\cline{8-11}
			\rowcolor{mygray}
			\multirow{-2}{*}{Method} & \multirow{-2}{*}{Pretrain} &  \multirow{-2}{*}{\shortstack{Training\\Frames}} & mAP & mAP$_s$ & mAP$_m$ & mAP$_l$ & Cls & Spat & Temp & Miss \\
			\hline
			\hline
			& ImageNet & & 29.27 & 6.17 & 27.88 & 33.50 & 8.12 & 8.13 & 9.87 & 3.73 \\
			\multirow{-2}{*}{Mask Track R-CNN} & MSCOCO & \multirow{-2}{*}{T=2} & 36.30 & 10.44 & 35.87 & 43.43 & 7.50 & 6.98 & 6.36 & 6.08 \\
			\hline
			& ImageNet & & 26.77 & 6.24 & 26.26 & 30.11 & 15.04 & 5.06 & 7.69 & 6.70 \\
			\multirow{-2}{*}{VISOLO} & MSCOCO & \multirow{-2}{*}{T=3} & 38.49 & 8.58& 39.31 & 40.94 & 12.35 & 2.74 & 8.97 & 5.84 \\																				
			\hline
			\hline
			& ImageNet & & 30.77 & 13.84 & 26.63 & 31.12 & 1.74 & 11.42 & 9.55 & 6.42 \\
			\multirow{-2}{*}{IFC} & MSCOCO & \multirow{-2}{*}{T=5} & 40.27 & 20.44 & 38.98 & 40.98 & 7.03 & 6.93 & 10.99 & 4.72 \\
			\hline
			& ImageNet & & 38.64 & 19.55 & 35.49 & 35.46 & 5.46 & 8.09 & 4.80 & 5.69 \\
			\multirow{-2}{*}{SeqFormer} & MSCOCO & \multirow{-2}{*}{T=5} & 43.37 & 21.69 & 40.43 & 39.20 & 6.95 &  7.69 & 4.00 & 6.20 \\
			\hline
			& ImageNet & & 35.14 & 19.84 & 33.05 & 34.24 & 5.14 & 6.19 & 6.81 & 5.65 \\
			\multirow{-2}{*}{Mask2Former} & MSCOCO & \multirow{-2}{*}{T=2} & 48.07 & 25.05 & 43.85 & 44.37 & 5.23 & 5.54 & 6.52 & 2.94 \\
			\hline
		\end{tabular}
	}
	\label{table:PretrainComparison}
\end{table*}
%+++++++++++++++++++

% table 4 ++++++++++
\begin{table*}[!t]
	\centering
	\captionsetup{font={normalsize}, justification=raggedright}
	\caption{\textbf{Performance cross models YouTube-VIS-2021 mini-val under different training frame numbers.} T represents the training frame number for each video. When more frames are involved during training, stronger temporal association ability reflected \textbf{TEMP} error weight decline.} 
	%%\vspace{3mm}
	\setlength\tabcolsep{3.2mm}
	\renewcommand\arraystretch{1.4}
	\scalebox{1.0}{
		\begin{tabular}{l|c|c|cccc|cccc}
			\hline\thickhline
			\rowcolor{mygray}
			& & & \multicolumn{4}{c|}{Metrics} & \multicolumn{4}{c}{Error Weights($\Delta$AP@50)} \\
			\cline{4-7}\cline{8-11}
			\rowcolor{mygray}
			\multirow{-2}{*}{Method} & \multirow{-2}{*}{Pretrain} &  \multirow{-2}{*}{\shortstack{Training\\Frames}} & mAP & mAP$_s$ & mAP$_m$ & mAP$_l$ & Cls & Spat & Temp & Miss \\
			\hline
			\hline
			& & T=3 & 38.49 & 8.58 & 39.31 & 40.94 & 12.35 & 2.74 & 8.97 & 5.84 \\
			\multirow{-2}{*}{VISOLO} & \multirow{-2}{*}{MSCOCO} & T=5 & 36.31 & 6.63 & 39.61 & 39.39 & 13.15 & 3.26 & 8.92 & 5.99 \\
			\hline
			\hline
			& & T=1 & 36.82 & 19.17 & 35.51 & 35.72 & 7.49 & 7.30 & 12.34 & 4.52 \\
			& & T=3 & 40.68 & 22.13 & 38.59 & 40.22 & 6.29 & 6.51 & 11.30 & 4.80 \\
			\multirow{-3}{*}{IFC} & \multirow{-3}{*}{MSCOCO} & T=5 & 40.27 & 20.44 & 38.98 & 40.98 & 7.03 & 6.93 & 10.99 & 4.72 \\
			\hline
			& & T=1 & 30.52 & 11.68 & 26.13 & 29.49 & 5.45 & 10.25 & 9.24 & 6.65 \\
			& & T=3 & 41.99 & 25.23 & 39.76 & 37.48 & 6.38 & 8.18 & 4.89 & 6.49 \\
			\multirow{-3}{*}{SeqFormer} & \multirow{-3}{*}{MSCOCO} & T=5 & 43.37 & 21.69 & 40.43 & 39.20 & 6.95 &  7.69 & 4.00 & 6.20 \\
			\hline
			& & T=1 & 45.41 & 26.87 & 42.21 & 42.86 & 3.23 & 5.39 & 11.40 & 3.62 \\ 
			& & T=3 & 47.89 & 24.42 & 43.86 & 44.56 & 5.92 & 6.33 & 5.97 & 3.18 \\
			\multirow{-3}{*}{Mask2Former} & \multirow{-3}{*}{MSCOCO} & T=5 & 47.83 & 25.01 & 43.97 & 44.46 & 3.30 & 5.85 & 6.07 & 3.76 \\
			\hline
		\end{tabular}
	}
	\label{table:FrameNumComparison}
	\vspace{-2mm}
\end{table*}
%+++++++++++++++++++

\textbf{Online methods.} As the pioneer algorithm, Mask Track R-CNN is built upon Mask R-CNN with an extra tracking branch. While VISOLO makes improvements by making full use of information from previous frames(\emph{e.g.}, category scores and mask features), under the guidance of grid similarity matching, formerly cues are aggregated to reweight scores and calibrate instance masks at the current frame. As shown in Figure.\ref{fig:OnlineModels}, with mask calibration, VISOLO has a much smaller weight of spatial segmentation error but achieves higher classification and temporal association error weights than Mask Track R-CNN, whose association is based on instance features within candidate boxes. This error distribution suggests that tracking by grid similarity and score reweighting are not as effective as the authors describe in their paper.

% figure 7 ++++++++++
\begin{figure*}[t]
	\centering
	\includegraphics[width=1.0\linewidth]{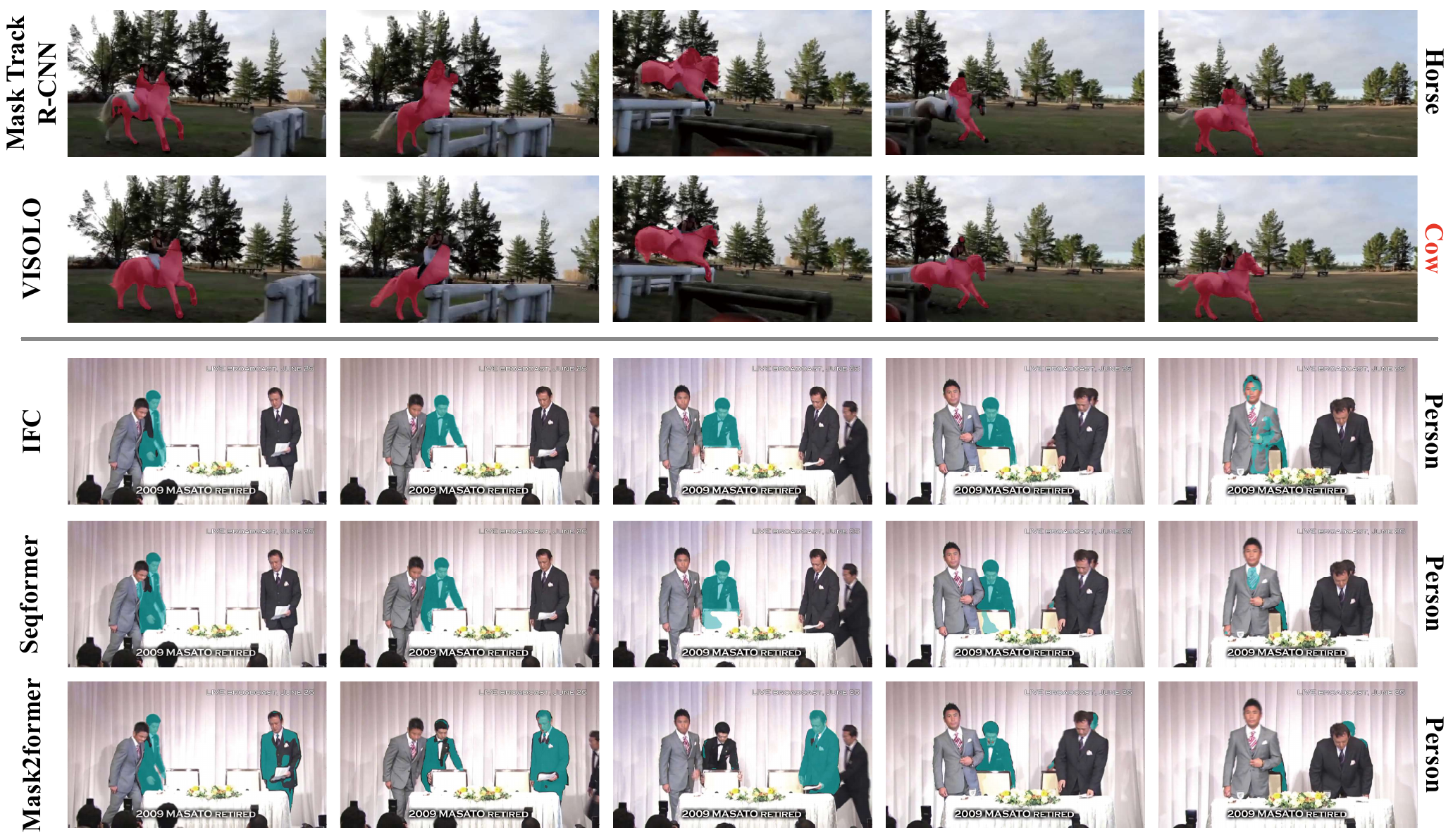}
	\captionsetup{font={normalsize}, justification=raggedright}
	\caption{\textbf{Qualitative comparisons cross models.} Visualization of predicted video instances in YouTube-VIS-2021 mini-val subset.}
	\label{fig:QualitativeComparison}
\end{figure*}
%+++++++++++++++++++

\vspace{0.3pt}
\textbf{Offline methods.} First, SeqFormer localizes objects by bounding boxes at each frame and aggregates box queries across the video to generate queries for video instances. Leveraging temporal context at instance-level, SeqFormer is naturally speculated to has a strong temporal association ability. Next, achieving inspiring performance on several image segmentation benchmarks\cite{Cocodataset,Cityscapes,Mapillary,Ade20k,Panoptic}, Mask2Former is extended to VIS task without any special modification. Its instance association is totally driven by the sequence loss criterion, so there is a strong belief that Mask2Former will outshine in spatial segmentation. As observed in Figure.\ref{fig:ErrorDist} and Figure.\ref{fig:OfflineModels}, SeqFormer unsurprisingly has the smallest temporal association error weight, and Mask2Former gains the least spatial segmentation errors. While IFC with global feature communication operation suffers from a much lower mAP, it surpasses the other two in both spatial and temporal localization errors.

%%%%% sec 4.2 %%%%%
\vspace{-1mm}
\subsection{Further analysis over attribute}
\label{sec:AttributeAnalysis}
In this subsection, we report models' performance over different instance temporal lengths. As illustrated in Table.\ref{table:MetricsOverTemporal}, all models fails to recognize over 70\% of the short video instances. While recognizing short video instances is especially difficult for online methods, their performance on the medium and long temporal range is on par with offline methods. We further report error distributions cross temporal ranges for VISOLO and Mask2Former for further investigation.

\vspace{1mm}
In Figure.\ref{fig:RangeErrorVISOLO}, VISOLO of the online inference paradigm is vulnerable to \textbf{Cls} and \textbf{Temp} error accumulation in the process of merging sequences when occlusion or fast motion exists, which can be even more severe in longer object sequences. And in Figure.\ref{fig:RangeErrorM2F}, Mask2Former gets progressively smaller temporal error weight as temporal length grows and results in a more balanced error proportion. This trend presents that Mask2Former is not good at localizing shorter video instances.

%%%%% sec 4.3 %%%%%
\subsection{Relation of spatial \& temporal localization} 
\label{sec:TemporalContext}
In this subsection, we explore how spatial segmentation and temporal association interact with each other.

We first investigate the impact of better spatial segmentation on temporal association. We replace the MSCOCO pretrain with ImageNet pretrain, thus should weaken the model's spatial segmentation ability, comparisons between the two pretrain mode are shown in Table.\ref{table:PretrainComparison}. When pretrained by instance segmentation task on MSCOCO, all methods achieves stronger spatial segmentation ability and much higher mAP. Mask Track R-CNN gets the most reduction in \textbf{Temp} error weight, SeqFormer and Mask2Former both have a light temporal association ability improvement, temporal association abilities of VISOLO and IFC even get harmed.

Then we change the training frame number to vary the model ability of temporal association. As shown in Table.\ref{table:FrameNumComparison}, when trained with more frames, all methods reduce their \textbf{Temp} error weights. More training frames bring unremarkable reductions in \textbf{Temp} error weights for VISOLO and IFC, spatial segmentation ability of IFC slightly benefits from better temporal association, but VISOLO suffers instead.  \textbf{Temp} error weights considerably decline for SeqFormer and Mask2Former, stronger temporal association of SeqFormer significantly promotes spatial segmentation, while Mask2Former's spatial segmentation ability stands independent of temporal association.

%%%%%%%%%% sec 4.4 %%%%%%%%%%
\subsection{Summary and Visualization}
As discussed in the three subsections above, offline inference algorithms surpass their online counterparts over all evaluation metrics. We find it is crucial for online methods to learn label consistency and location variation as time flows. Frame-level query aggregation can significantly help the offline model to alleviate temporal miss-association. Except for SeqFormer with frame-level query aggregation, there is no other approach that enables spatial segmentation and temporal association to benefit from each other. 

Besides quantitive experiments, we select two video instances and group the visualization into online and offline paradigms, qualitative comparisons of selected algorithms are shown in Figure.\ref{fig:QualitativeComparison}. As observed, VISOLO miss-classifies the horse while segments better than Mask Track R-CNN. For offline methods, IFC balances the qualities between spatial segmentation and temporal association, SeqFormer successfully tracked the targeted person, but with extra pixels assigned to its adjacent objects, Mask2Former is puzzled in distinguishing object instances with large similarity.

%%%%%%%%%% sec 5 %%%%%%%%%%
\section{Conclusion}
In this work, we propose a novel error analyzing toolbox for VIS, which defines meaningful error types with a focus on spatial-temporal localization quality. By weighting errors, we successfully indicate model discrepancies, and we report model performance over instance temporal length. Extensive experiments show that most investigated algorithms cannot leverage the model abilities of spatial segmentation and temporal association. They generally aim at promoting one aspect. We expect our proposed toolbox can give a clear picture of model characters in modeling video instances at pixel-level, and give interpretable suggestions for algorithm design. The proposed toolbox can be easily extended to support other video object recognition tasks, which may be the future work of the toolbox.



\bibliographystyle{elsarticle-num} 
\bibliography{TIVE-Refs}


%% \begin{thebibliography}{00}


%% \bibitem{}

%% \end{thebibliography}

\end{document}